\documentclass[letterpaper]{article}
\usepackage{aaai20}
\usepackage{times}
\usepackage{helvet}
\usepackage{courier}
\usepackage[hyphens]{url} 
\usepackage{graphicx}
\usepackage{graphicx}
\usepackage{amsmath}
\usepackage{amssymb}

\usepackage{multirow}
\usepackage{bm}
\usepackage{comment}
\usepackage{algorithm}
\usepackage{algorithmic}
\usepackage{xcolor}

\newcommand{\keypoint}[1]{\vspace{0.1cm}\noindent\textbf{#1}}
\newcommand{\cut}[1]{}
\newcommand{\tableCellHeight}{1.1}
\title{Deep Domain-Adversarial Image Generation for Domain Generalisation}
\author{
Kaiyang Zhou\textsuperscript{\rm 1}, Yongxin Yang\textsuperscript{\rm 1}, Timothy Hospedales\textsuperscript{\rm 2,3}, and Tao Xiang\textsuperscript{\rm 1,3} \\
\textsuperscript{\rm 1}University of Surrey, \textsuperscript{\rm 2}University of Edinburgh, \textsuperscript{\rm 3}Samsung AI Center, Cambridge \\
\{k.zhou, yongxin.yang, t.xiang\}@surrey.ac.uk, t.hospedales@ed.ac.uk
}

\begin{document}

\maketitle
\begin{abstract}
Machine learning models typically suffer from the domain shift problem when trained on a source dataset and evaluated on a target dataset of different distribution. To overcome this problem, domain generalisation (DG) methods aim to leverage data from multiple source domains so that a trained model can generalise to unseen domains. In this paper, we propose a novel DG approach based on \emph{Deep Domain-Adversarial Image Generation} (DDAIG). Specifically, DDAIG consists of three components, namely a label classifier, a domain classifier and a domain transformation network (DoTNet). The goal for DoTNet is to map the source training data to unseen domains. This is achieved by having a learning objective formulated to ensure that the generated data can be correctly classified by the label classifier while fooling the domain classifier. By augmenting the source training data with the generated unseen domain data, we can make the label classifier more robust to unknown domain changes. Extensive experiments on four DG datasets demonstrate the effectiveness of our approach.
\end{abstract}

\section{Introduction} \label{sec:intro}
\begin{figure}[t]
    \centering
    \includegraphics[width=\columnwidth]{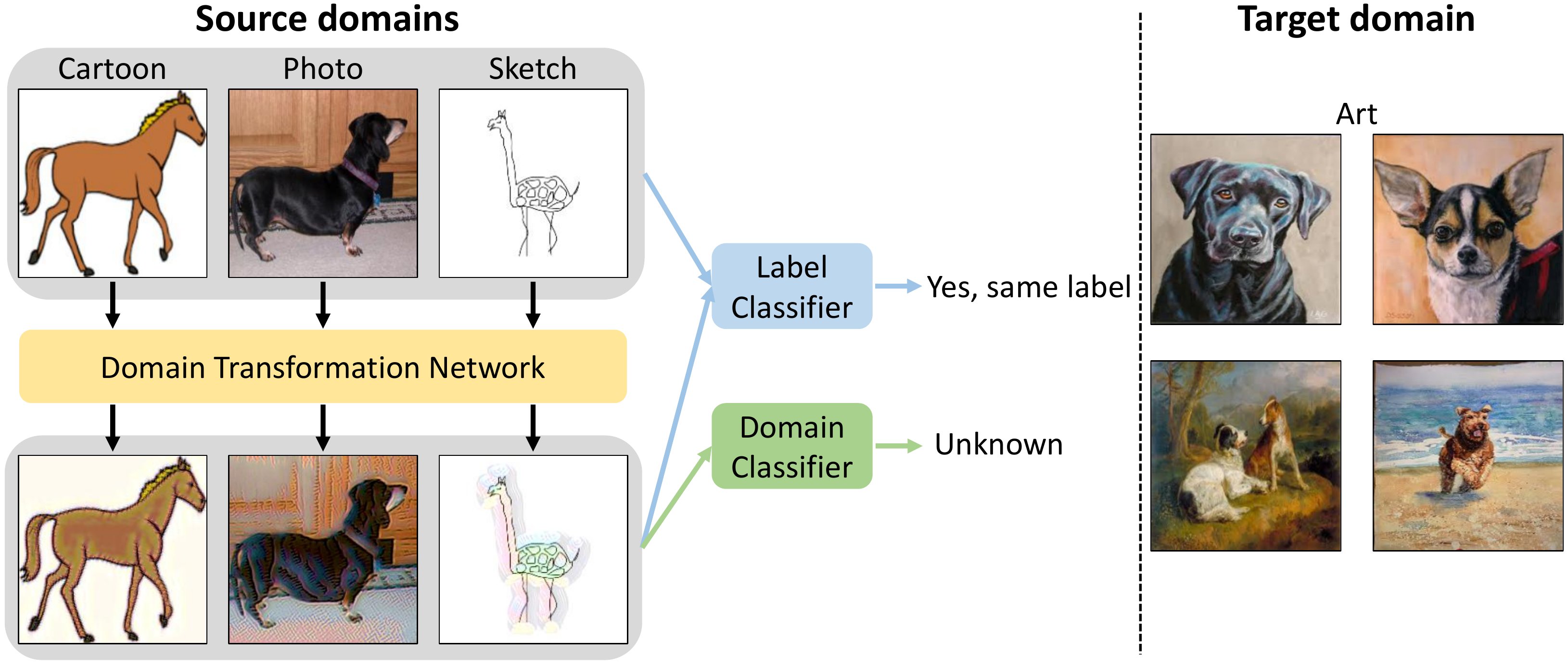}
    \caption{\small Given multiple source domains, e.g. Cartoon, Photo and Sketch, we learn a domain transformation network (DoTNet) to transform images to unseen domains, which maintain the class labels but change domain-related properties. Both original and transformed images are used to train a label classifier, which is applied to an unseen target domain, e.g. Art.}
    \label{fig:concept_illustration}
    \vspace{-0.2cm}
\end{figure}

Most existing deep learning models assume that the training (source) and testing (target) data come from the same domain/dataset and thus follow the same distribution. However, in practice this assumption is often invalid. For example, a module for recognising pedestrians  and traffic signs in an autonomous driving car may be deployed anywhere in the world under any weather condition. Considering each city and weather combination as a domain, it is impossible to collect training data of every domain for model training. Other domain changes can correspond to the change of image style/modality such as those shown in Figure~\ref{fig:concept_illustration} where a classifier trained on images of cartoon, photo and sketch is applied to art images. Unfortunately, it is well known that existing deep learning models are sensitive to domain changes/shifts~\cite{li2017deeper,shankar2018generalizing,balaji2018metareg} in that they tend to overfit the source domains, resulting in poor generalisation.

A straightforward way to deal with the domain gap between source and target domains is to acquire labelled target data and perform supervised model fine-tuning. However, large-scale data collection and annotation for every new target domain is prohibitively expensive and time-consuming, which make the fine-tuning strategy infeasible. A more economical solution is to use unsupervised domain adaptation (UDA) methods~\cite{long2015learning,ganin2015unsupervised,hoffman2018cycada,cvpr19dlow,chen2019trans}, which only use unlabelled target data. Although the data annotation step is avoided, UDA still requires a data collection step followed by a model adaptation step for each new domain, which hinders its applicability.

As a result, domain generalisation (DG)~\cite{muandet2013domain} has received an increasing interest lately. The goal of DG is to train a model using data from multiple source domains and deploy the model to an arbitrary unseen target domain without any adaption. Many existing DG methods adopt a core idea from the domain adaption (DA) research, which is to align source domain distributions at feature-level, assuming that a source domain invariant model can be learned~\cite{li2018mmdaae,li2018ciddg}. However, without access to any target domain data, the model learned with domain alignment can still overfit the source domains. Alternatively, meta learning based methods have been recently employed to address DG where held-out source domains are used to simulate unseen target domains~\cite{li2018learning,balaji2018metareg}. However, meta learning models still focus on narrowing domain gaps among source domains and thus offer no guarantee for generalisation to unseen domains.

In this paper, we tackle the DG problem by \emph{synthesising data from unseen domains}. We assume that augmenting the original training data of source domains with synthetic data from unseen domains could make the task model intrinsically more domain-generalisable~\cite{DomainRandomization,yue2019domain}. To this end, a novel framework based on \emph{Deep Domain-Adversarial Image Generation} (DDAIG) is introduced, which is illustrated in Figure~\ref{fig:concept_illustration}. There are three components in DDAIG, which are label classifier, domain classifier and domain transformation network (DoTNet). Each component is a deep neural network. The label classifier and domain classifier are trained to predict the class labels and domain labels of the input data respectively. The functionality of DoTNet is to transform the input data in such a way that they can be recognised by the label classifier but fool the domain classifier. In particular, the transformation produced by DoTNet is designed to be perturbations with the same shape as the input. Therefore, the new data is generated by combining the perturbations with the original input. By doing so, we can efficiently generate additional training data that covers the (otherwise sparsely sampled) manifold of domains, which in turn allows a more domain-agnostic label classifier to be learned. We further show that DoTNet can be easily extended to incorporate other types of transformations, e.g., geometric transformations by adding spatial transformer network (STN)~\cite{nips15stn}. In practice, the three networks are trained jointly in an end-to-end manner. Unlike the domain alignment and meta learning methods, our DDAIG works directly at pixel-level, thus largely improving the interpretability of the model.

To evaluate DDAIG, we conduct extensive experiments on three DG benchmark datasets, namely PACS~\cite{li2017deeper}, Office-Home~\cite{office_home} and digit recognition among MNIST~\cite{lecun1998mnist}, MNIST-M~\cite{ganin2015unsupervised}, SVHN~\cite{netzer2011svhn} and SYN~\cite{ganin2015unsupervised}. These datasets cover a variety of visual recognition tasks and contain different types of domain variation (see Figure~\ref{fig:example_images}). We demonstrate that DDAIG outperforms current state-of-the-art DG methods on all datasets. We also verify the effectiveness of DDAIG on a heterogeneous DG task, i.e., person re-identification where the source and target domains have different label spaces. Finally, we visualise the generated images and feature embeddings to provide insights on why our approach works.
The code is available at {https://github.com/KaiyangZhou/DG-research-pytorch}.

\begin{figure*}[ht]
\centering
\includegraphics[width=.85\textwidth]{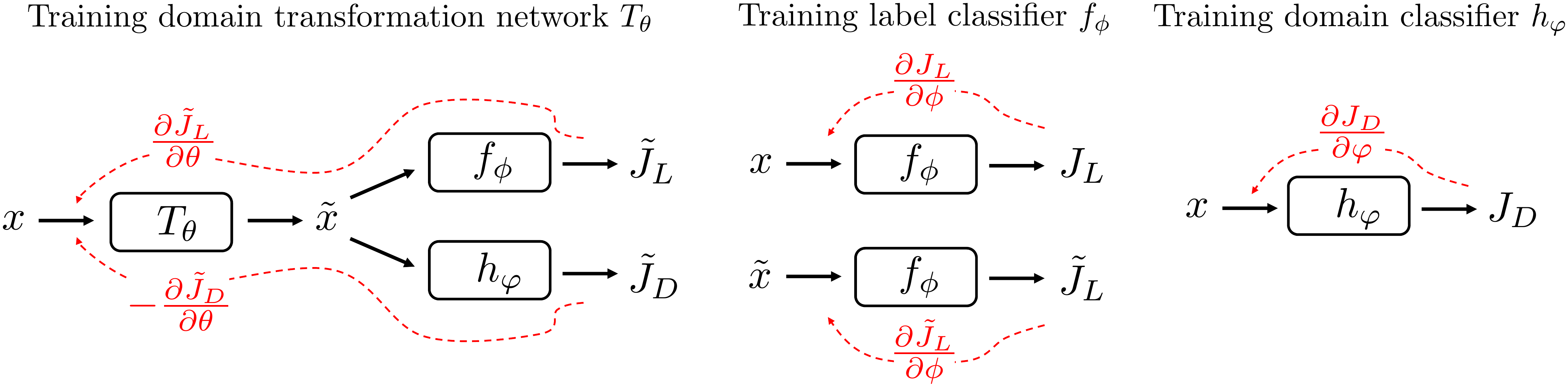}
\caption{\small Overview of our framework. A domain transformation network $T_\theta$ is trained by minimising the label classification loss $\tilde{J}_L$ while maximising the domain classification loss $\tilde{J}_D$ on the transformed data $\tilde{x}$. The label classifier $f_\phi$ is learned by minimising the label classification loss given both original and transformed data. The domain classifier $h_\varphi$ is trained to classify each instance into one of source domains. The red dashed arrows represent the gradient flow.}
\label{fig:framework_overview}
\vspace{-0.2cm}
\end{figure*}

\section{Related Work} \label{sec:related_work}
Early kernel alignment~\cite{muandet2013domain,gan2016learning} and examplar SVM based~\cite{xu2014exploiting,niu2015multiview} DG models have been followed mainly by deep neural network based ones. The current deep DG studies can be generally divided into three groups: (i) domain alignment, (ii) meta learning and (iii) data augmentation.

\keypoint{Domain alignment} has been extensively studied for domain adaptation (DA) problems, where some labelled or unlabelled data are accessible during training. These DA methods aim to either (i) minimise the distance (e.g., maximum mean discrepancy (MMD)~\cite{pan2008transfer}) between source and target distributions, or (ii) fool a domain classifier that tries to discriminate different domains~\cite{ganin2015unsupervised}. As shown in \cite{motiian2017unified}, domain alignment based DA models can be easily modified for DG by iteratively training on every pair of source domains. Among the recent alignment based deep DG models, \cite{li2018mmdaae} proposed to minimise MMD of all possible pairs within source domains, meanwhile an adversarial autoencoder was used to ensure that the learned features follow the Laplace distribution. \cite{li2018ciddg} considered aligning the conditional distributions as well as the marginal ones via adversarial training.
Though domain alignment is a sensible strategy for DA, the potential risk of applying it to DG is that the model might overfit all seen domains yet still generalise poorly to the unseen domains.

\keypoint{Meta learning} in computer vision has been widely exploited for few-shot learning~\cite{finn2017model}. Recently, meta learning has been adapted to address the DG setting~\cite{li2018learning,balaji2018metareg}. Since the final objective of a DG model is to generalise to unseen domains, the key idea of using meta learning is to simulate domain shift during training to prepare models for domain shift during testing. Specifically, source domains are separated into two disjoint sets, namely meta-train and meta-validation, and a model is optimised on meta-train so as to boost the performance on meta-validation. One early work in this direction is MLDG~\cite{li2018learning}, which is based on MAML~\cite{finn2017model}. Recently, MetaReg~\cite{balaji2018metareg} proposed to learn a customised regulariser to improve DG. A meta learning based DG approach is appealing as it reduces the efforts in manual design. However, as a black-box approach it is hard to diagnose exactly how it improves the DG performance. Importantly, using only the original source domain data, it still has the risk of overfitting source domains.

\begin{figure}[t]
    \centering
    \includegraphics[width=.8\columnwidth]{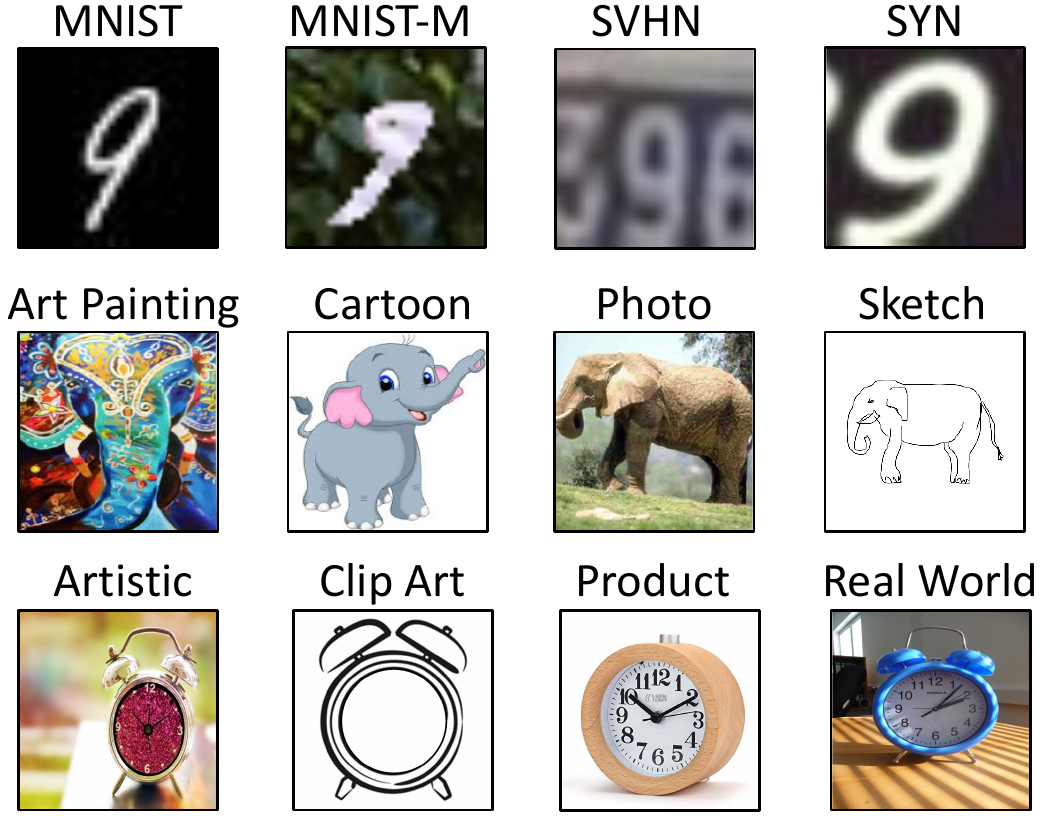}
    \caption{\small 
    Example images from Digits-DG (1st row), PACS (2nd row) and Office-Home (3rd row) show that large domain gaps exist, challenging domain generalisation.
    }
    \label{fig:example_images}
    \vspace{-0.4cm}
\end{figure}

\keypoint{Data augmentation} is a common practice to train deep neural networks, e.g. flipping and rotation. However, conventional data augmentation methods only deal with simple geometric changes within the same dataset~\cite{volpi2019model}. When the domain gap is large such as those illustrated in Figure~\ref{fig:example_images} containing image style variations, learning-based augmentation strategies are required. Very recently, inspired by adversarial attacks~\cite{goodfellow2015explaining}, \cite{shankar2018generalizing} introduced CrossGrad to generate a new sample $\tilde{x}$ by adding to the original sample its gradient from a domain classifier $h(\cdot)$, i.e., $\tilde{x} \leftarrow x + \epsilon \frac{\partial h(x)}{\partial x}$. A similar approach with an additional regularisation term in $h(\cdot)$ was proposed in~\cite{volpi2018generalizing} for the single source domain case. The main drawback of these methods is their direct and simple dependence on the gradient, which only makes simple perturbations that cannot account for semantic changes like style or font shift (see Figure~\ref{fig:vs_crossgrad}). Moreover, being based on adversarial attack models that are deliberately designed to make imperceptible modifications to an image, the perturbations are too subtle to be representative of real-world domain shift. In contrast, by \emph{learning} a full CNN model (i.e. DoTNet) to generate the `shift', we can produce more sophisticated and more overt perturbations to synthesise new data and the results are easier to interpret. We demonstrate clear advantages of our approach over CrossGrad both quantitatively and qualitatively. We also show that our transformation CNN can be easily extended to incorporate geometric transformations such as rotation (see Figure~\ref{fig:vis_stn_ptb}), which is impossible with gradient-based perturbation methods.

\vspace{-0.1cm}
\section{Methodology} \label{sec:methodology}
Our idea to tackle domain generalisation (DG) is based on \emph{Deep Domain-Adversarial Image Generation}~(DDAIG), which aims to train a domain transformation network (DoTNet) to synthesise data from unseen domains given some input and use both original and synthetic data to learn a domain-invariant classifier. To learn DoTNet, we simultaneously train a label classifier and a domain classifier, which are tasked to recognise the class labels and domains of the input data, respectively. The learning objective for DoTNet is to transform the input data in such a way that the synthetic data keeps the same labels as the input but fools the domain classifier. As the synthetic data has labels, it can be combined with the original data to train the label classifier using supervised learning. As a result, the label classifier can learn representations that are more invariant to domain shift than that trained with the original data only\footnote{To clarify, a classifier means the combination of a feature extraction backbone and a softmax classification layer, unless specified otherwise.}. An overview of our DDAIG framework is illustrated in Figure~\ref{fig:framework_overview}. The learning procedure for each component is detailed below.

\keypoint{Domain transformation network}.
Let $T_\theta$ be the DoTNet parameterised by $\theta$, $f_\phi$ the label classifier parameterised by $\phi$, $h_\varphi$ the domain classifier parameterised by $\varphi$, $y$ the class label of input $x$ and $d$ the domain label, the objective function for $T_\theta$ is
\begin{equation} \label{eq:obj_T}
\min_\theta~ \tilde{J}_L (f_\phi (T_\theta(x)), y) - \tilde{J}_D (h_\varphi (T_\theta(x)), d),
\end{equation}
where $\tilde{J}_L$ and $\tilde{J}_D$ are cross-entropy losses for label and domain classification, respectively. We use differentiable neural networks to construct $T_\theta$, $f_\phi$ and $h_\varphi$, thus the gradients can be back-propagated through $f_\phi$ and $h_\varphi$ and all the way to $T_\theta$. The specific architecture  design of $T_\theta$ will be discussed later.

\keypoint{Label classifier}.
The label classifier $f_\phi$ is fed with both original and synthetic data. The loss function is
\begin{equation} \label{eq:obj_L}
\min_\phi~ (1 - \alpha) J_L (f_\phi(x), y) + \alpha \tilde{J}_L (f_\phi (T_\theta(x)), y),
\end{equation}
where $\alpha$ is a balance weight, which is fixed to 0.5.

\keypoint{Domain classifier}.
The domain classifier $h_\varphi$ is required to capture domain-discriminative features, thus its learning objective is to minimise the domain classification loss w.r.t $\varphi$,
\begin{equation} \label{eq:obj_D}
\min_\varphi~ J_D (h_\varphi (x), d).
\end{equation}

Note that our domain classifier is analogous to the discriminator in the classic GAN framework~\cite{goodfellow2014generative} but differs in that we do multi-class classification~\cite{odena2017conditional} (on source domains\footnote{In DG tasks, we often assume that we have access to multiple source domains.}) while GAN's discriminator performs binary classification (real or fake). This difference ensures that maximising the domain classification loss does not simply force the synthetic data to fall into another single domain distribution. Suppose there are three source domains, given a synthetic instance from the first domain we maximise $- \log \frac{ e^{z_1} }{ e^{z_1} + e^{z_2} + e^{z_3} }$ ($z_i$ denotes logit), which essentially minimises $z_1$ whilst giving \emph{equal gradients} to maximise $z_2$ and $z_3$ simultaneously. As such, neither one of the gradients to $z_2$ and $z_3$ is dominant.

\keypoint{Architecture design}.
For the label classifier, any network architecture suitable for the given recognition problem can be adopted. Throughout this paper, the domain and label classifiers share the same architecture. To construct DoTNet, we use fully convolutional network (FCN)~\cite{seg_fcn}. Instead of directly generating data, which is often difficult because the data can be high-dimensional such as RGB image, we use FCN to generate perturbations, which are added to the input, resulting in
\begin{equation} \label{eq:design_T}
\tilde{x} = x + \lambda T_\theta(x),
\end{equation}
where $\lambda$ is a positive weight typically set between 0.1 and 0.7. This is inspired by the residual feature learning~\cite{he2016deep} but the residual connection links the input directly to the output. This design is also related to adversarial attack methods~\cite{szegedy2014intriguing,goodfellow2015explaining}.
However, different from adversarial perturbations, which are usually imperceptible, our DoTNet is allowed to produce visually perceptible perturbations, which can better represent the real-world domain shift (see Figure~\ref{fig:vis_ptb}). Note that (\ref{eq:design_T}) will replace the $T_\theta(x)$ in (\ref{eq:obj_T}) and (\ref{eq:obj_L}).

\begin{figure}[t]
    \centering
    \includegraphics[width=.95\columnwidth]{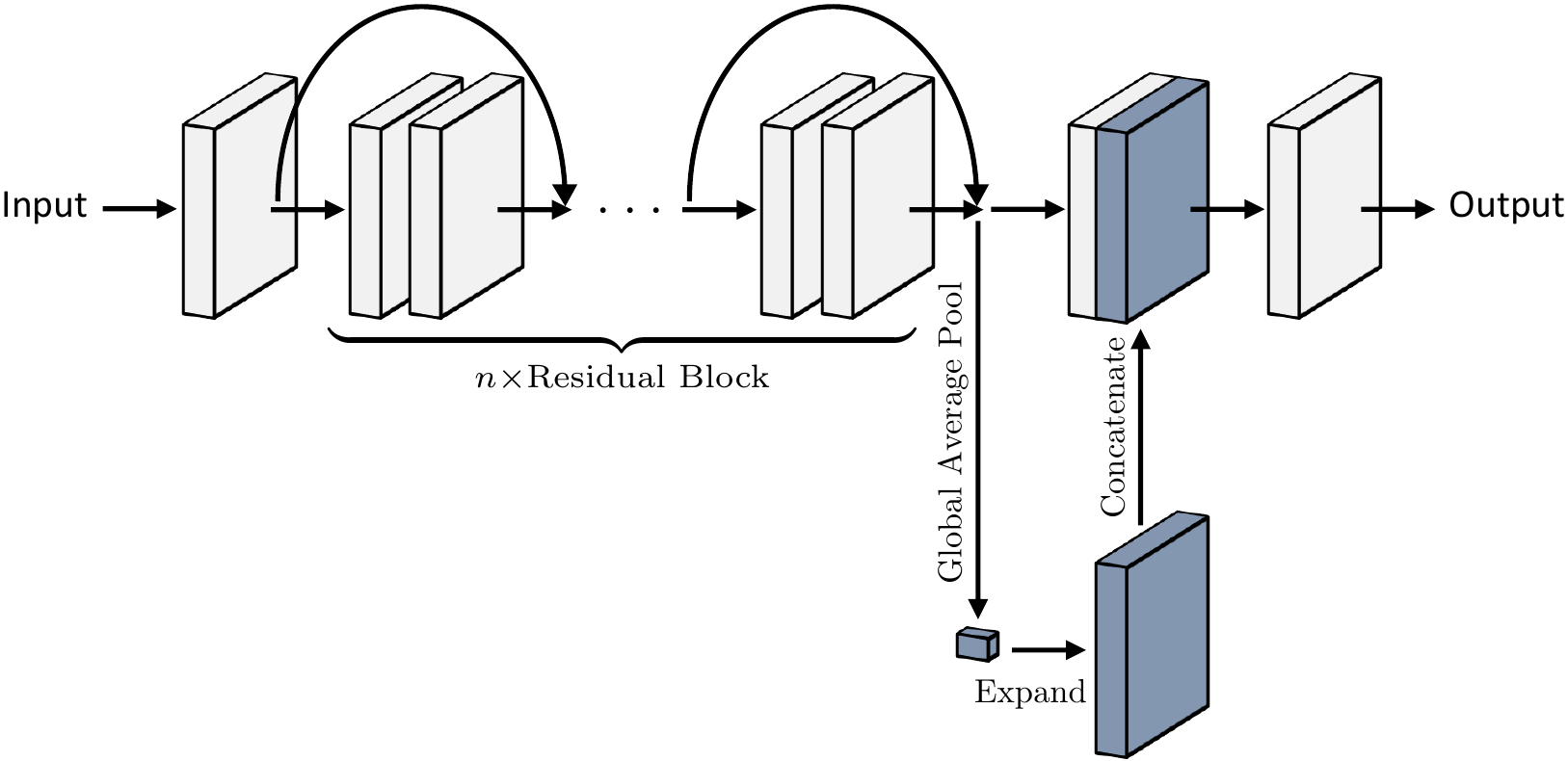}
    \caption{\small Architecture of domain transformation network.}
    \label{fig:arch_T}
    \vspace{-0.4cm}
\end{figure}

More concretely, as shown in Figure~\ref{fig:arch_T}, the architecture of DoTNet starts with a $3 \times 3$ conv layer, followed by $n$ 2-conv residual blocks~\cite{he2016deep} to extract mid-level features. All residual blocks use $3 \times 3$ kernels. The output is branched into an identity layer and a global average-pooling layer. The latter produces a context vector encapsulating the global information~\cite{liu2016parsenet}. The context vector is expanded spatially and then concatenated back to the main stream, which is further processed by a $1 \times 1$ conv layer for feature fusion. Finally, a $1 \times 1$ conv layer is used to generate the perturbation. All layers use ReLU as the non-linearity function except the last one which uses the Tanh function. Similar to~\cite{zhu2017unpaired}, we insert instance normalisation layer~\cite{Ulyanov_2017_CVPR} after every conv layer excluding the last one. The stride is set to 1 for all layers. All $3 \times 3$ kernels use the reflection padding of size 1.

In this paper, we mainly investigate this perturbation design for $T$. However, our framework is generic and does not impose any restriction on the architecture  (as long as it is differentiable). The design of $T$ thus mostly depends on the specific tasks at hand.
For instance, $T$ can take the form of STN~\cite{nips15stn} to deal with geometric transformations; or even a combination of STN and the perturbation architecture in  Figure~\ref{fig:arch_T}.

The full algorithm of DDAIG is presented in Algorithm~\ref{alg:our_method}. Note that the warm-up scheme mainly aims to make the data generated by DoTNet more reliable before being fed to the classifier.

\begin{algorithm}[t]
   \caption{Deep Domain-Adversarial Image Generation}
   \label{alg:our_method}
   \footnotesize
\begin{algorithmic}[1] 
   \STATE {\bf Input:} source domains $\mathcal{S}$, label classifier $f_\phi$, domain classifier $h_\varphi$, DoTNet $T_\theta$, learning rate $\eta$, hyperparameter $\lambda$, loss balance weight $\alpha$, maximum iteration $K$, warmup iteration $K_\mathrm{m}$.
   \STATE {\bf Output:} label classifier $f_\phi$.
   \FOR{$k = 1$ {\bf to} $K$}
     \STATE $(x, y, d) \sim \mathcal{S}$ \hfill // \textit{Randomly sample a minibatch}
     \STATE $\tilde{x} = x + \lambda T_\theta(x)$ \hfill // \textit{Transform the minibatch}
     \STATE $\theta = \theta - \eta \nabla_\theta (\tilde{J}_L - \tilde{J}_D)$ \hfill // \textit{Update DoTNet}
     \IF{$k < K_\mathrm{m}$}
         \STATE $\phi = \phi - \eta \nabla_\phi J_L$ \hfill // \textit{Update label classifier}
     \ELSE
         \STATE $\tilde{x} = x + \lambda T_\theta(x)$ \hfill // \textit{Transform the minibatch using updated DoTNet}
         \STATE $\phi = \phi - \eta \nabla_\phi ((1 - \alpha) J_L + \alpha \tilde{J}_L$) \hfill // \textit{Update label classifier using both original and synthetic data}
     \ENDIF
     \STATE $\varphi = \varphi - \eta \nabla_\varphi J_D$ \hfill // \textit{Update domain classifier}
   \ENDFOR
\end{algorithmic}
\end{algorithm}

\section{Experiments} \label{sec:experiments}

\subsection{Datasets and Settings}
We first evaluate our approach DDAIG on three conventional DG benchmark datasets, which cover a variety of recognition problems.
(1) We conduct leave-one-domain-out digit recognition on \emph{MNIST}~\cite{lecun1998mnist}, \emph{MNIST-M}~\cite{ganin2015unsupervised}, \emph{SVHN}~\cite{netzer2011svhn} and \emph{SYN}~\cite{ganin2015unsupervised}, which differ drastically in font style and background (see Figure~\ref{fig:example_images} 1st row). MNIST contains handwritten digit images. MNIST-M is a variant of MNIST by blending the images with random colour patches. SVHN contains street view house number images. SYN consists of synthetic digit images with varying fonts, backgrounds and stroke colours. This benchmark is called \textbf{Digits-DG} hereafter. We randomly select 600 images per class from each dataset and split the data into 80\% for training and 20\% for validation. All models are trained on the training data of three domains and evaluated on all images of the remaining domain.
(2) \textbf{PACS}~\cite{li2017deeper} consists of four domains, namely \emph{Photo} (1,670 images), \emph{Art Painting} (2,048 images), \emph{Cartoon} (2,344 images) and \emph{Sketch} (3,929 images). Each domain contains seven categories. Following the prior work~\cite{li2017deeper,cvpr19JiGen,li2019episodic}, we choose one domain as test domain and use the remaining three domains as source domains. To fairly compare with published methods, our models are trained using data only from the training split. The domain shift mainly corresponds to image style changes as depicted in Figure~\ref{fig:example_images} 2nd row.
(3) \textbf{Office-Home}~\cite{office_home}, originally introduced for domain adaptation, is getting popular in the DG community~\cite{d2018domain,cvpr19JiGen}. It contains four domains, which are \emph{Artistic}, \emph{Clipart}, \emph{Product} and \emph{Real World}. Each domain has 65 classes, which are related to office and home objects. There are around 15,500 images in total. The domain variations mainly take place in background, viewpoint and image style. See Figure~\ref{fig:example_images} (the third row) for example images.

For performance measure, we report top-1 classification accuracy (\%) averaged over five runs and 95\% confidence intervals. We compare our DDAIG with state-of-the-art DG methods with reported results on these datasets or codes. These methods include CCSA~\cite{motiian2017unified}, MMD-AAE~\cite{li2018mmdaae}, CrossGrad~\cite{shankar2018generalizing}, MetaReg~\cite{balaji2018metareg}, D-SAM~\cite{d2018domain}, JiGen~\cite{cvpr19JiGen} and Epi-FCR~\cite{li2019episodic}. We also include a strong baseline called Vanilla, which directly combines data from all source domains for model training without any DG-targeting tricks.

\begin{table}[t]
\setlength{\tabcolsep}{3pt}
\renewcommand{\arraystretch}{\tableCellHeight}
\centering
\footnotesize
\caption{Leave-one-domain-out results on Digits-DG dataset (with 95\% confidence intervals).}
\label{tab:resOnDigitsDG}
\begin{tabular}{l | c c c c | c}
\hline
Method & MNIST & MNIST-M & SVHN & SYN & Avg. \\ \hline
Vanilla & 95.8$\pm$.3 & 58.8$\pm$.5 & 61.7$\pm$.5 & 78.6$\pm$.6 & 73.7 \\
CCSA & 95.2$\pm$.2 & 58.2$\pm$.6 & {65.5}$\pm$.2 & 79.1$\pm$.8 & 74.5 \\
MMD-AAE & 96.5$\pm$.1 & 58.4$\pm$.1 & 65.0$\pm$.1 & 78.4$\pm$.2 & 74.6 \\
CrossGrad & \textbf{96.7}$\pm$.1 & {61.1}$\pm$.5 & 65.3$\pm$.5 & {80.2}$\pm$.2 & {75.8} \\
DDAIG (\emph{ours)} & {96.6}$\pm$.2 & \textbf{64.1}$\pm$.4 & \textbf{68.6}$\pm$.6 & \textbf{81.0}$\pm$.5 & \textbf{77.6} \\
\hline
\end{tabular}
\vspace{-0.3cm}
\end{table}

\subsection{Evaluation on Digits-DG}
\keypoint{Implementation}.
Images are resized to $32\times32$ and converted to RGB by replicating channels.
The classification network is constructed by four $3 \times 3$ conv layers (64 kernels), each followed by ReLU and $2 \times 2$ max-pooling. A softmax classification layer is attached on the top, which takes the flattened vector as input.
The networks are trained from scratch using SGD, initial learning rate of 0.05, batch size of 128 and weight decay of 5e-4 for 50 epochs. The learning rate is decayed by 0.1 every 20 epochs.

\keypoint{Results}.
Table~\ref{tab:resOnDigitsDG} shows that our DDAIG achieves the best overall performance (Avg.), outperforming the second best CrossGrad by a clear margin of 1.8\% and all domain alignment methods (CCSA \& MMD-AAE) by more than 3\%. On the most difficult target domains, namely MNIST-M and SVHN which contain complex backgrounds and cluttered digits respectively, DDAIG obtains large margins over the competitors, notably with +5.3\% and +6.9\% improvements compared with the Vanilla model. This demonstrates the effectiveness of the generated unseen domain data, which essentially increases the diversity of source domains.

\subsection{Evaluation on PACS}
\keypoint{Implementation}.
Images are resized to $224 \times 224$. Following~\cite{cvpr19JiGen,li2019episodic}, we use the ImageNet-pretrained ResNet18~\cite{he2016deep} as the classification network.
The networks are trained with SGD, initial learning rate of 5e-4, batch size of 16 and weight decay of 5e-4 for 25 epochs. The learning rate is decayed by 0.1 at the 20th epoch.
For data augmentation, we use random crop on images rescaled by a factor of 1.25 and random horizontal flip.
During the first three epochs, the label classifier is only fed with real data.

\keypoint{Results}.
We summarise our findings from Table~\ref{tab:resOnPACS} as follows.
(1) Our DDAIG is clearly the best  method, beating the second best methods MetaReg and Epi-FCR by around 1.5\% (on Avg.). It is noted that MetaReg benefits from additional training data from the validation split. The recently proposed Epi-FCR uses episodic training to improve the Vanilla model. DDAIG outperforms Epi-FCR on all domains by a clear margin, suggesting that data augmentation with unseen domain data is much more effective.
(2) Again, DDAIG achieves large improvements (3.7\%+ on Avg.) against all domain alignment methods.
(3) Compared with CrossGrad, DDAIG yields large improvements on all domains except Photo. In particular, the margins are 4.4\% on Art, 1.3\% on Cartoon and 4.5\% on Sketch. This justifies that learning a dedicated CNN for perturbation generation is much more useful than gradient-based perturbations.

\begin{table}[t]
\setlength{\tabcolsep}{4pt}
\renewcommand{\arraystretch}{\tableCellHeight}
\centering
\footnotesize
\caption{Leave-one-domain-out results on PACS dataset (with 95\% confidence intervals). $\dagger$: results are reported in their papers. $\ddagger$: use train+val for training.}
\label{tab:resOnPACS}
\begin{tabular}{l | c c c c | c}
\hline
Method & Art & Cartoon & Photo & Sketch & Avg. \\ \hline
MetaReg$^{\dagger\ddagger}$ & 83.7$\pm$.1 & 77.2$\pm$.3 & 95.5$\pm$.2 & 70.3$\pm$.3 & 81.7 \\ \hline
Vanilla & 77.0$\pm$.6 & 75.9$\pm$.6 & \textbf{96.0}$\pm$.1 & 69.2$\pm$.6 & 79.5  \\
CCSA & 80.5$\pm$.6 & 76.9$\pm$.6 & 93.6$\pm$.4 & 66.8$\pm$.9 & 79.4 \\
MMD-AAE & 75.2$\pm$.3 & 72.7$\pm$.3 & \textbf{96.0}$\pm$.1 & 64.2$\pm$.2 & 77.0 \\
CrossGrad & 79.8$\pm$.8 & 76.8$\pm$.8 & \textbf{96.0}$\pm$.2 & 70.2$\pm$.4 & 80.7 \\
D-SAM$^\dagger$ & 77.3 & 72.4 & {95.3} & \textbf{77.8} & 80.7 \\
JiGen$^\dagger$ & 79.4 & 75.3 & \textbf{96.0} & 71.6 & 80.5 \\
Epi-FCR$^\dagger$ & {82.1} & {77.0} & 93.9 & 73.0 & {81.5} \\
DDAIG (\emph{ours)}) & \textbf{84.2}$\pm$.3 & \textbf{78.1}$\pm$.6 & {95.3}$\pm$.4 & {74.7}$\pm$.8 & \textbf{83.1} \\
\hline
\end{tabular}
\vspace{-0.1cm}
\end{table}

\begin{table}[t]
\setlength{\tabcolsep}{3pt}
\renewcommand{\arraystretch}{\tableCellHeight}
\centering
\footnotesize
\caption{Leave-one-domain-out results on Office-Home dataset (with 95\% confidence intervals). $\dagger$: results are reported in their papers.}
\label{tab:resOnOfficeHome}
\begin{tabular}{l | c c c c | c}
\hline
Method & Artistic & Clipart & Product & Real World & Avg. \\ \hline
Vanilla & 58.9$\pm$.3 & 49.4$\pm$.1 & {74.3}$\pm$.1 & \textbf{76.2}$\pm$.2 & 64.7  \\
CCSA & \textbf{59.9}$\pm$.3 & {49.9}$\pm$.4 & 74.1$\pm$.2 & 75.7$\pm$.2 & {64.9} \\
MMD-AAE & 56.5$\pm$.4 & 47.3$\pm$.3 & 72.1$\pm$.3 & 74.8$\pm$.2 & 62.7  \\
CrossGrad & 58.4$\pm$.7 & 49.4$\pm$.4 & 73.9$\pm$.2 & 75.8$\pm$.1 & 64.4 \\
D-SAM$^\dagger$ & 58.0 & 44.4 & 69.2 & 71.5 & 60.8 \\
JiGen$^\dagger$ & 53.0 & 47.5 & 71.5 & 72.8 & 61.2 \\
DDAIG (\emph{ours)}) & {59.2}$\pm$.1 & \textbf{52.3}$\pm$.3 & \textbf{74.6}$\pm$.3 & {76.0}$\pm$.1 & \textbf{65.5} \\
\hline
\end{tabular}
\vspace{-0.4cm}
\end{table}

\subsection{Evaluation on Office-Home}

\keypoint{Implementation}.
Following~\cite{d2018domain,cvpr19JiGen}, we randomly split the data into 90\% for training and 10\% for validation. The commonly used leave-one-domain-out protocol is adopted for evaluation. For fair comparison with published methods, we only use the training split of source domains for model training. Other implementation details for network training are the same as those for PACS.

\keypoint{Results}.
From Table~\ref{tab:resOnOfficeHome}, we observe that the Vanilla model achieves very strong performance, largely outperforming most DG methods including MMD-AAE, D-SAM and JiGen. This makes sense because the domain gap is much smaller compared to that in PACS, especially among Artistic, Product and Real World, where the variations mainly take place in background and viewpoint. Among all methods, only DDAIG achieves a clear margin against Vanilla. In particular, it is worth noting that DDAIG achieves huge improvement (+2.9\%) on Clipart, which contains the largest domain gap as opposed to the source domains (see Figure~\ref{fig:example_images}). Therefore, the results strongly demonstrate the versatility of our DDAIG framework. Compared with CrossGrad, DDAIG achieves better performance on all domains.

\begin{table}[t]
\setlength{\tabcolsep}{2.8pt}
\renewcommand{\arraystretch}{\tableCellHeight}
\centering
\footnotesize
\caption{Results on cross-domain person re-ID datasets.}
\label{tab:xdomainreid}
\begin{tabular}{l | c c c c | c c c c}
\hline
\multirow{2}{*}{Method} & \multicolumn{4}{c|}{Market1501$\rightarrow$Duke} & \multicolumn{4}{c}{Duke$\rightarrow$Market1501} \\
 & R1 & R5 & R10 & mAP & R1 & R5 & R10 & mAP \\
\hline
Vanilla & 48.5 & 62.3 & 67.4 & 26.7 & {57.7} & 73.7 & {80.0} & 26.1 \\
CrossGrad & 48.5 & 63.5 & {69.5} & 27.1 & 56.7 & 73.5 & 79.5 & {26.3} \\
DDAIG (\emph{ours}) & \textbf{50.6} & \textbf{65.2} & \textbf{70.3} & \textbf{28.6} & \textbf{60.9} & \textbf{77.1} & \textbf{83.2} & \textbf{29.0} \\
\hline
\end{tabular}
\vspace{-0.1cm}
\end{table}

\begin{table}[t]
\setlength{\tabcolsep}{3.5pt}
\renewcommand{\arraystretch}{\tableCellHeight}
\centering
\footnotesize
\caption{Results on Digits-DG using different $\lambda$'s.}
\label{tab:impactOfLmda}
\begin{tabular}{l | c c c c | c}
\hline
Method & MNIST & MNIST-M & SVHN & SYN & Avg. \\
\hline
Baseline & 95.8 & 58.8 & 61.7 & 78.6 & 73.7 \\
DDAIG $\lambda=0.1$ & 96.3 & 62.3 & 68.6 & 79.8 & 76.8 \\
DDAIG $\lambda=0.3$ & 96.4 & 61.9 & 68.0 & 81.0 & 76.8 \\
DDAIG $\lambda=0.5$ & 96.6 & 61.2 & 68.0 & 80.5 & 76.6 \\
DDAIG $\lambda=0.7$ & 96.4 & 64.1 & 65.9 & 80.8 & 76.8 \\
\hline
\end{tabular}
\vspace{-0.3cm}
\end{table}

\subsection{Evaluation on Heterogeneous DG}
In this section, we evaluate our approach on the cross-dataset person re-identification (re-ID) task, which is essentially a heterogeneous DG problem due to disjoint label spaces between training and test identities~\cite{feature_critic}. Person re-ID aims to match people across non-overlapping camera views. In this task, we treat each camera view as a domain.

\keypoint{Datasets}.
We use two commonly used re-ID datasets, namely \emph{Market1501}~\cite{zheng2015scalable} and \emph{DukeMTMC-reID} (Duke)~\cite{ristani2016perform,zheng2017unlabeled}. Market1501 contains 32,668 images of 1,501 identities, which are captured by 6 cameras. Duke contains 36,411 images of 1,812 identities, which are captured by 8 cameras. Each dataset is split into training set, query set and gallery set based on the standard protocols~\cite{zheng2015scalable,zheng2017unlabeled}. For evaluation, we train models using one dataset and perform test on the other. Cumulative Matching Characteristics (CMC) ranks and mean Average Precision (mAP) are used as the performance measure.

\keypoint{Implementation}.
Images are resized to $256 \times 128$. We adopt the state-of-the-art re-ID model OSNet~\cite{zhou2019osnet,zhou2019learning} as the CNN backbone. Following~\cite{zhou2019osnet,zhou2019learning}, we train the re-ID model using the standard classification pipeline where each person identity is regarded as a class. Therefore, the entire training algorithm remains the same as before. At test time, the features extracted from the re-ID model are used to compute Euclidean distance for image matching. The code is based on Torchreid~\cite{torchreid}.

\keypoint{Results}.
We compare our method with CrossGrad and the strong vanilla model.
The overall results are shown in Table~\ref{tab:xdomainreid}. It is clear that only our DDAIG consistently improves upon the vanilla baseline on both settings, with noticeable margins.
It is widely acknowledged that cross-domain re-ID is a challenging problem~\cite{zhong2019invariance,zhou2019learning}. Without using target data, it is difficult to gain improvement over the vanilla model. Therefore, the results strongly demonstrate the versatility of our DDAIG: It is not only effective for the conventional DG tasks but also useful to heterogeneous DG problems such as person re-ID.

\begin{table}[t]
    \setlength{\tabcolsep}{7pt}
    \renewcommand{\arraystretch}{\tableCellHeight}
    \centering
    \footnotesize
    \caption{Accuracy on rotated MNIST when MNIST-M, SVHN and SYN are used as the source domains.}
    \label{tab:rotatedMNIST}
    \begin{tabular}{l | c c c c c}
    \hline
    Method & $0^\circ$ & $20^\circ$ & $30^\circ$ & $40^\circ$ & $50^\circ$ \\
    \hline
    CrossGrad & 96.7 & 82.7 & 65.1 & 45.6 & 30.1 \\
    DDAIG \emph{w/o STN} & 96.6 & 87.7 & 72.1 & 54.1 & 40.3 \\
    DDAIG \emph{w/ STN} & 96.4 & 87.7 & 76.7 & 61.1 & 46.6 \\
    \hline
    \end{tabular}
    \vspace{-0.2cm}
\end{table}

\begin{figure}[t]
\centering
\includegraphics[width=.9\columnwidth]{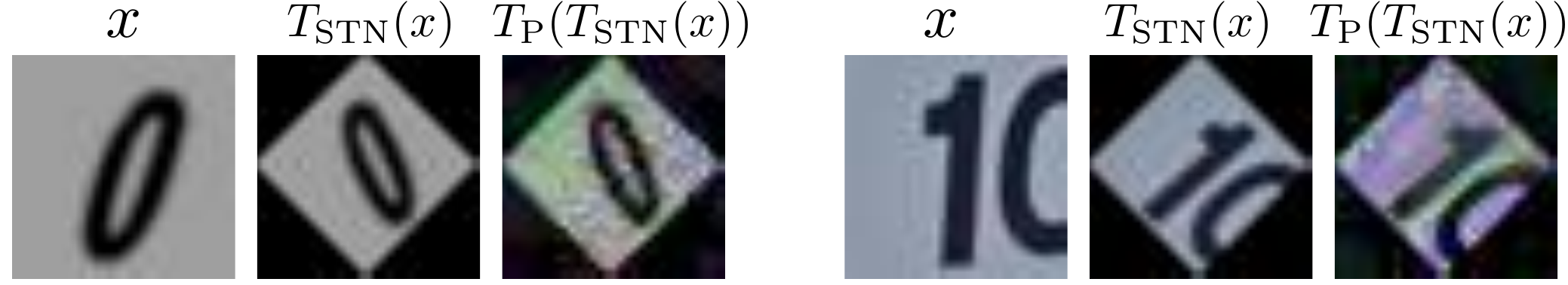}
\caption{\small Transformations produced by an extended DoTNet. STN: Spatial Transformer Network. P: Perturbation.}
\label{fig:vis_stn_ptb}
\vspace{-0.3cm}
\end{figure}

\begin{table}[t]
\setlength{\tabcolsep}{3.5pt}
\renewcommand{\arraystretch}{\tableCellHeight}
\centering
\footnotesize
\caption{Comparison between models trained using source data, novel data and source+novel data.}
\label{tab:vs_newDataonly}
\begin{tabular}{c c | c c c c c}
\hline
Souce & Novel & MNIST & MNIST-M & SVHN & SYN & Avg. \\
\hline
$\checkmark$ & & 95.8 & 58.8 & 61.7 & 78.6 & 73.7 \\ 
 & $\checkmark$ & 95.6 & 58.3 & 57.9 & 79.9 & 72.9 \\
$\checkmark$ & $\checkmark$ & 96.6 & 64.1 & 68.6 & 81.0 & 77.6 \\
\hline
\end{tabular}
\vspace{-0.3cm}
\end{table}

\begin{figure*}[t]
\centering
\includegraphics[width=0.95\textwidth]{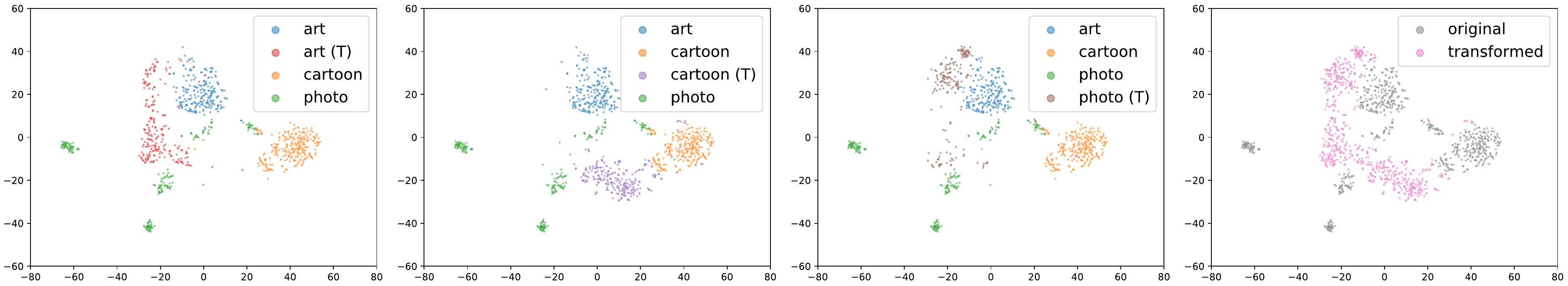}
\caption{\small T-SNE visualisation in the domain space using PACS's validation set. The first three images compare transformed data (T) with original data for each source domain. The last image shows an overall comparison between original (grey) and transformed (pink) data.}
\label{fig:vis_dom_embed_pacs}
\vspace{-0.1cm}
\end{figure*}

\subsection{Further Analysis}
\keypoint{Impact of $\lambda$}.
Table~\ref{tab:impactOfLmda} shows the results of varying $\lambda$ in Eq.~\ref{eq:design_T} from 0.1 to 0.7. When the target domain is less dissimilar to the source domains such as MNIST and SYN, the result does not vary too much with different $\lambda$s. However, when the target domain has a larger domain gap than the sources, e.g. MNIST-M and SVHN, our model shows a moderate sensitivity to $\lambda$. It is important to note that all results are better than the Vanilla baseline.

\keypoint{Dealing with geometric transformations}.
Image perturbation is less useful in simulating geometric transformations such as rotation. To overcome this limitation, we extend the perturbation CNN (Figure~\ref{fig:arch_T}) by inserting STN~\cite{nips15stn} before it. Therefore, the new DoTNet first geometrically transforms the input and then adds perturbation. Note that such a transformation is  impossible with gradient-based perturbation methods~\cite{shankar2018generalizing}.
We test this design on Digits-DG where the target domain is rotated MNIST and the source domains are MNIST-M, SVHN and SYN. Note that the `rotation' shift is never observed among the sources.
The results are shown in Table~\ref{tab:rotatedMNIST}. First of all, we observe that all methods' performance drops as the rotation degree increases. This is expected because increasing the rotation degree essentially enlarges the domain gap with the source data, making the target domain more challenging.
Comparing DDAIG (without STN) with CrossGrad, the performance drop of the latter is much larger. This indicates that the CNN-learned perturbation does not contain solely the style changes but also sophisticated transformations, which make the task model more robust to the geometric domain shift.
With STN, the performance drop is significantly reduced (especially on $40^\circ$ and $50^\circ$) which demonstrates the flexibility of the DDAIG framework.
See Figure~\ref{fig:vis_stn_ptb} for the visualisation of the transformations.

\keypoint{Importance of combining source and novel data}.
Table~\ref{tab:vs_newDataonly} shows that training with the novel data only does not bring any gain at all. This is expected because the performance gain of DDAIG mainly comes from the aggregation of source domains and the generated novel domains.

\begin{figure}[t]
\centering
\includegraphics[width=.75\columnwidth]{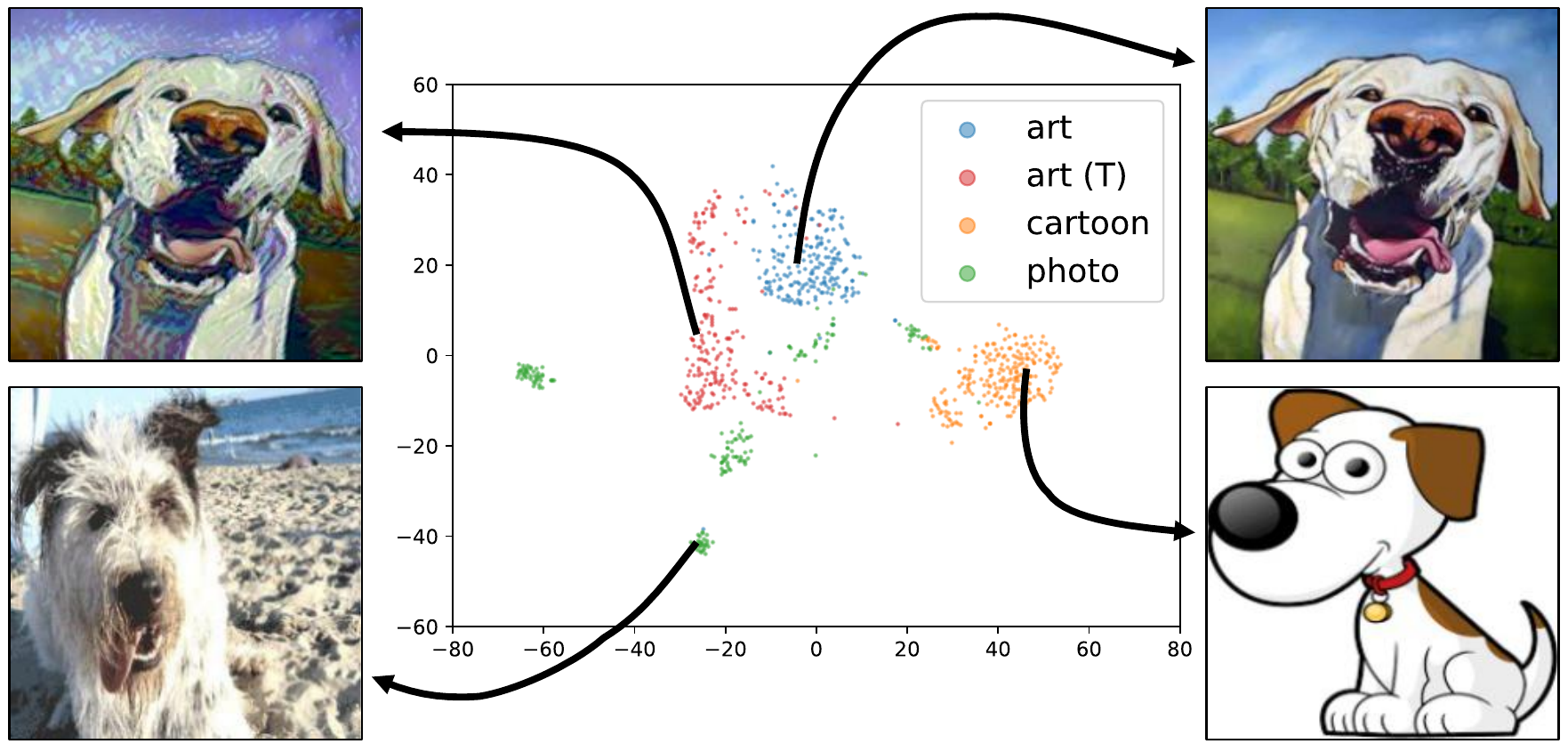}
\caption{\small Transformed image vs. original source images.}
\label{fig:vis_dspace_and_image}
\vspace{-0.2cm}
\end{figure}

\begin{figure}[t]
\centering
\includegraphics[width=.95\columnwidth]{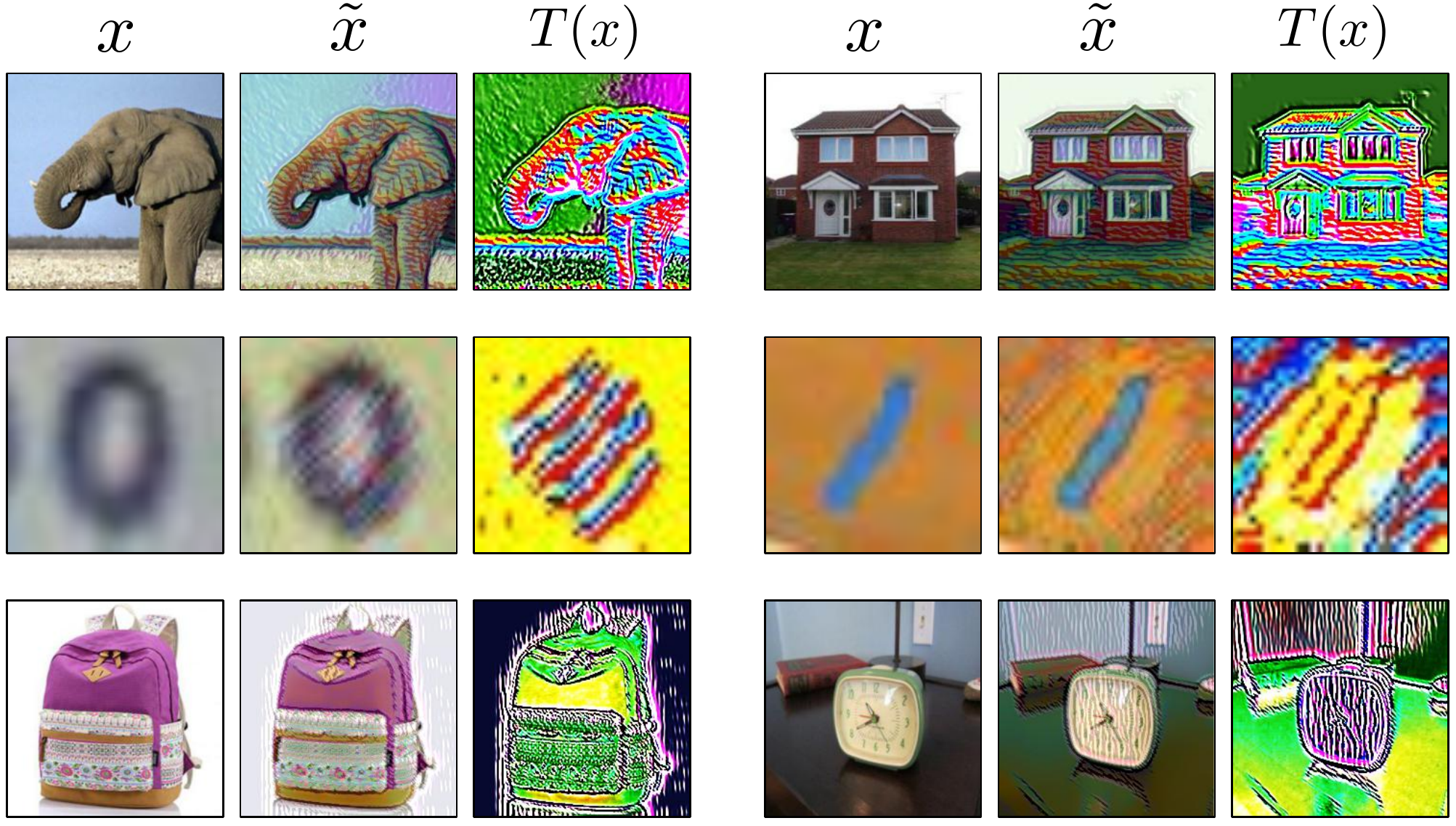}
\caption{\small Examples of transformed images from PACS (1st row), Digits-DG (2nd row) and Office-Home (3rd row). $x$, $\tilde{x}$ and $T(x)$ denote original image, transformed image and transformation (perturbation), respectively.}
\label{fig:vis_ptb}
\vspace{-0.3cm}
\end{figure}

\begin{figure}[t]
\centering
\includegraphics[width=.98\columnwidth]{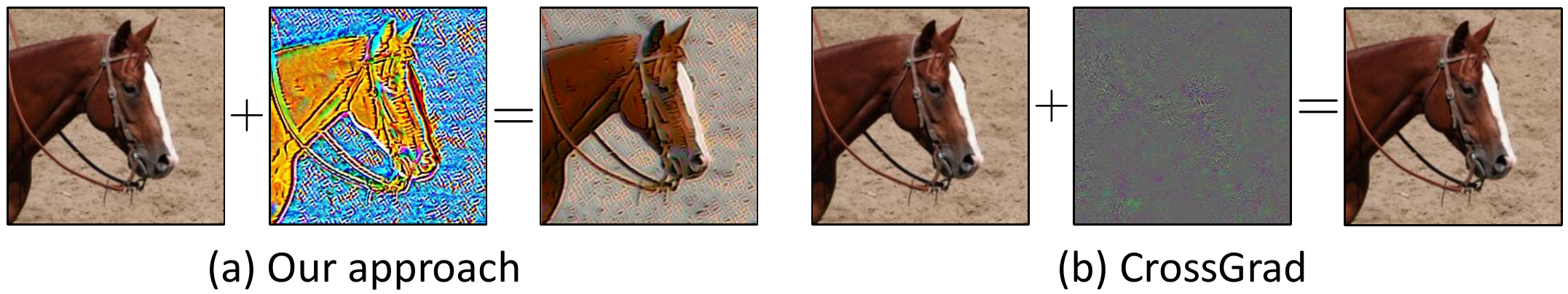}
\caption{\small Comparison between perturbations generated by CNN (ours) and adversarial perturbations.}
\label{fig:vs_crossgrad}
\vspace{-0.2cm}
\end{figure}

\keypoint{Visualisation}.
To better understand why DDAIG works for DG, we visualise the feature embeddings in the domain space using t-SNE~\cite{tsne} (see Figure~\ref{fig:vis_dom_embed_pacs}). It is clear that the new data distributions do not overlap with any of the existing source domains. Instead, they are distributed over the unfilled domain space, indicating exploration of unseen domains. Consequently, the generated unseen-domain data along with the source-domain data allows the model to learn more domain-generalisable representations, which explains why DDAIG achieves excellent performance on all DG benchmark datasets. Figure~\ref{fig:vis_dspace_and_image} shows four example images from the first domain space where the new ``art'' image clearly differs from the other source images in terms of image style.

Figure~\ref{fig:vis_ptb} provides a clearer view of how images are transformed by the perturbations. Comparing the transformed images with the original images, we observe that the domain-related information has been drastically changed while the category-specific properties are well maintained, which is consistent with the motivation of our method. The perturbations are instance-specific and can represent complex transformations such as colour and texture. For instance, the perturbation for the elephant image tends to add green and pink colours to the background and enhance the textures on the elephant body.

\keypoint{Comparison with adversarial perturbations}.
Figure~\ref{fig:vs_crossgrad} compares the perturbations between DDAIG and CrossGrad. It is obvious that CrossGrad's perturbation does not contain meaningful patterns and look like salt-and-pepper noise, resembling those of adversarial attack methods~\cite{goodfellow2014generative}. In contrast, our perturbation is instance-specific and has obvious effects on the transformed image, which is more representative of the real-world domain shift.

\section{Conclusion} \label{sec:conclusion}
We presented DDAIG, a novel DG method to synthesise data from unseen domains for data augmentation. Unlike current data augmentation-based DG methods, DDAIG learns a full transformation CNN to model the domain shift. Extensive experiments on three DG datasets showed that our method can improve the generalisation of CNN models on unseen domains, outperforming current state-of-the-art DG methods. Results on the cross-domain person re-ID task further demonstrated the versatility of DDAIG beyond DG.

{\small
\bibliography{reference}
\bibliographystyle{aaai}
}

\end{document}